\documentclass[10pt,twocolumn,letterpaper]{article}

\usepackage[pagenumbers]{wacv} 

\usepackage{graphicx}
\usepackage{amsmath}
\usepackage{amssymb}
\usepackage{booktabs}
\usepackage{float}

\usepackage[pagebackref,breaklinks,colorlinks]{hyperref}

\usepackage[capitalize]{cleveref}
\crefname{section}{Sec.}{Secs.}
\Crefname{section}{Section}{Sections}
\Crefname{table}{Table}{Tables}
\crefname{table}{Tab.}{Tabs.}

\def\wacvPaperID{1910} 
\def\confName{WACV}
\def\confYear{2025}

\newcommand{\mycomment}[1]{}

\begin{document}

\title{SymFace: Additional Facial Symmetry Loss for Deep Face Recognition}

\author{Pritesh Prakash\\
Central Research Laboratory\\
  Bharat Electronics Limited\\
  Ghaziabad, India, 201010 \\
{\tt\small priteshprakash@bel.co.in}
\and
Koteswar Rao Jerripothula\\
  Department of Electrical Engineering\\
  IIT Kanpur\\
  India, 208016 \\
{\tt\small kotesrj@iitk.ac.in}
\and
Ashish Jacob Sam\\
  Central Research Laboratory\\
  Bharat Electronics Limited\\
  Ghaziabad, India, 201010 \\
{\tt\small ashishjacobsam@bel.co.in}
\and
Prinsh Kumar Singh\\
  Central Research Laboratory\\
  Bharat Electronics Limited\\
  Ghaziabad, India, 201010 \\
{\tt\small prinshkumarsingh@bel.co.in}
\and
S Umamaheswaran\\
  Central Research Laboratory\\
  Bharat Electronics Limited\\
  Ghaziabad, India, 201010 \\
{\tt\small sumamaheswaran@bel.co.in}
}
 \maketitle

\begin{abstract}
  Over the past decade, there has been a steady advancement in enhancing face recognition algorithms leveraging advanced machine learning methods. The role of the loss function is pivotal in addressing face verification problems and playing a game-changing role. These loss functions have mainly explored variations among intra-class or inter-class separation. This research examines the natural phenomenon of facial symmetry in the face verification problem. The symmetry between the left and right hemi faces has been widely used in many research areas in recent decades. This paper adopts this simple approach judiciously by splitting the face image vertically into two halves. With the assumption that the natural phenomena of facial symmetry can enhance face verification methodology, we hypothesize that the two output embedding vectors of split faces must project close to each other in the output embedding space. Inspired by this concept, we penalize the network based on the disparity of embedding of the symmetrical pair of split faces. Symmetrical loss has the potential to minimize minor asymmetric features due to facial expression and lightning conditions, hence significantly increasing the inter-class variance among the classes and leading to more reliable face embedding. This loss function propels any network to outperform its baseline performance across all existing network architectures and configurations, enabling us to achieve SoTA results.
\end{abstract}

\section{Introduction}
\label{sec:intro}

The symmetry between the left and right hemi faces is a natural phenomenon. Absolute symmetry between the two sides of faces can be rare, but it's also uncommon to find someone with a highly asymmetrical face. The degree of symmetry varies from person to person; some may possess a highly symmetrical face, while others may not. Facial asymmetry can stem from various factors, including genetic predispositions, developmental irregularities, traumatic incidents, or other influences impacting the formation and growth of facial structures. These factors can result in visible differences in the positioning and proportions of facial features such as the eyes, nose, mouth, and ears, which are often noticeable in cases of facial asymmetry. The study of face symmetry holds a significant backbone in many research domains. Researchers usually utilize facial symmetry as a metric for assessing attractiveness \cite{FaceSymAttractNature, FaceSymAttract2}, gauging emotional expressions \cite{SymEmotion}, investigating neurological disorders \cite{symNeuro} and Deepfake analysis \cite{FaceSymDeepFake}; its applications span across fields like psychology, anthropology, and medicine, offering insights into various aspects of human biology, behavior, and perception etc. 

\begin{figure}[t]
    \centering
    \begin{subfigure}[b]{0.4\linewidth}
        \centering
        \fbox{\includegraphics[width=0.8\linewidth]{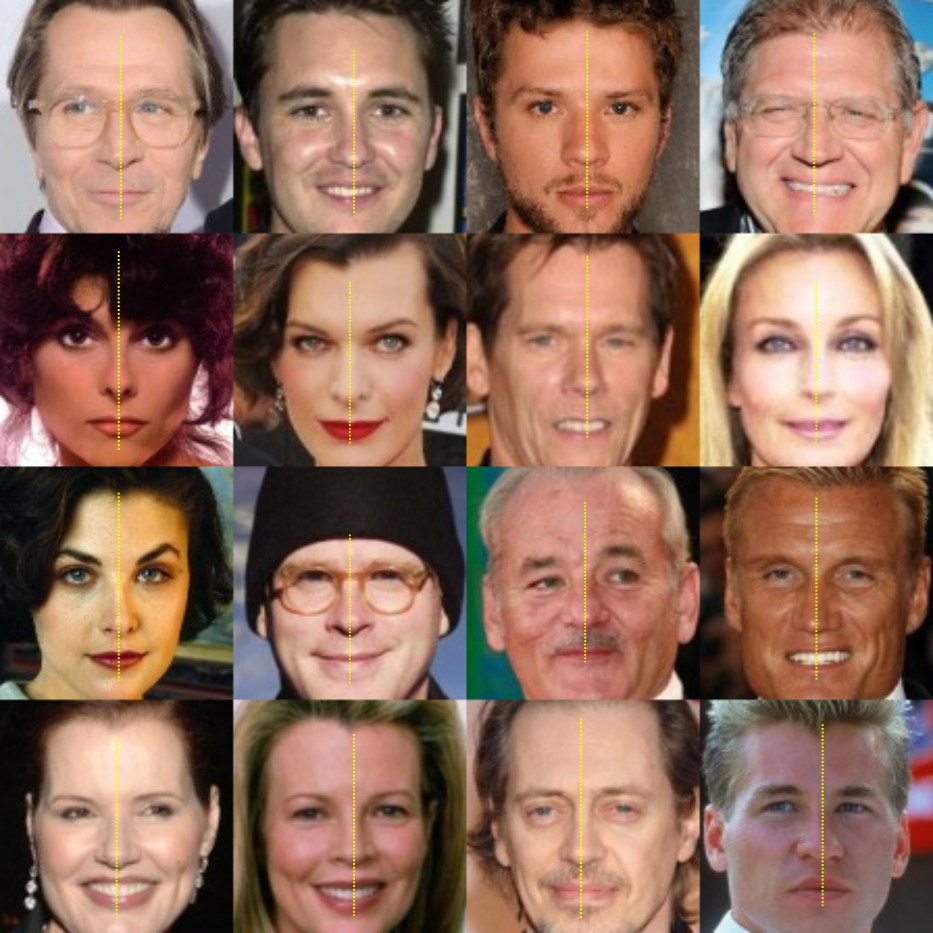}}
        \caption{Samples with good symmetrical features}
        \label{fig-high_soc}
    \end{subfigure}
    \begin{subfigure}[b]{0.4\linewidth}
        \centering
        \fbox{\includegraphics[width=0.8\linewidth]{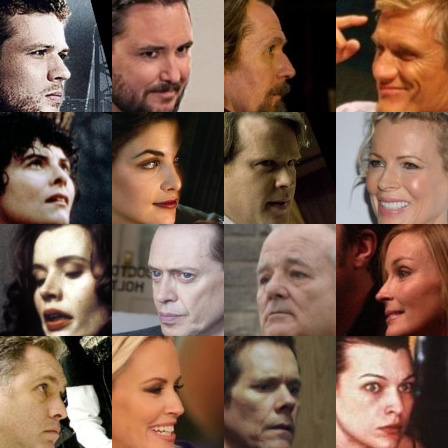}}
        \caption{Samples with poor symmetrical features}
        \label{fig-low_soc}
    \end{subfigure}
    \caption{Symmetrical analysis on face dataset}
    \label{fig:whole}
\end{figure}

Due to the association of symmetrical behaviour between the left and right hemi faces, we applied the symmetrical influence in the existing face recognition methods. Measuring symmetry in face data is only feasible if the camera is positioned appropriately in front of the face. Although projecting from 3D to 2D loses some symmetrical aspects, there is still enough symmetrical information retained in the 2D images from the analysis. The available face datasets \cite{casia, MS-celeb-1M} consist of face images captured from various angles and positions (Fig. \ref{fig:whole}), causing a significant variance in the view angle and orientation of the face. This is a natural occurrence but imposes a real hindrance in existing face recognition datasets and may affect the ability of the network to learn the symmetrical factors in the face recognition problem. If we analyze the faces at extreme view angles, it is observed that the mind barely recognizes the symmetry between the hemi faces. In such cases, only one hemi face is visible. Therefore, we do not derive the symmetrical factors from such images, which are tilted more than a marginal angle. However, we feed these images to the network without adding the symmetrical aspects. This approach allows us to apply the symmetrical effect only to well-oriented face images. This way, the network learns to extract the asymmetric features more precisely from the image and outperforms the benchmarks in side pose datasets \cite{CFP-FP, CP-LFW}.

We've developed a standard technique called the 3-Point Symmetric Split (3PSS) algorithm for assessing and assigning symmetric orientation coefficient ($\rho$) in facial features. A high value of $\rho$ implies a good orientation for detecting symmetry, and a lower value of $\rho$ implies a poor orientation for detecting symmetry in facial features. The 3PSS algorithm is designed to analyze symmetrical orientation in 2D space. Consequently, images may get classified as lower $\rho$ because of the face orientation in the 2D space, despite the individual having a naturally symmetrical face. While 3PSS provides valuable insights into facial symmetry, its application can be limited to specific types of research, and it is advised to avoid erroneous conclusions. The 3PSS algorithm categorizes each image of datasets as symmetrical or asymmetrical with a corresponding $\rho$ value.

In the past decade, various novel approaches have been explored to enhance the discriminating power of the network and demonstrate outstanding results in the face recognition domain. The main principle behind the previous research is to increase inter-class and reduce the intra-class variation among the classes. The network output, i.e., the positioning of the vector embedding in the embedding space of two input images of the same class, should be projected close to each other. So, we hypothesize that the vertically split faces of the same person belonging to the same input will also be projected much closer to each other in the embedding space. With this hypothesis, we introduce an innovative method for integrating SymFace loss. The network can be trained to minimize the distance between any complete facial feature and hemi faces belonging to the same class. Adding the SymFace loss with the standard face losses achieves SoTA results in various networks and surpasses the existing benchmark datasets \cite{LFW, AgeDB, CA-LFW, CP-LFW, CFP-FP} in the face recognition domain.

The key feature of the proposed method can be summarized as:
\begin{itemize}
\item We introduce the influence of facial symmetry in the face recognition domain. In this proposed methodology, we define a systematic approach to apply SymFace loss from the data augmentation to the loss calculation.

\item We propose a method that navigates the 2-D space, significantly reducing manual effort and computational overhead in exploring symmetry. However, this method is not recommended for measuring symmetry in the face for any generic purposes.

\item We propose a theory that the vertically split front 2-D face image possesses the property of symmetry, and two symmetrical halves should be close enough to each other in the output embedding space, implying that the L2 distance between the embedding of the two halves of such hemi faces should be minimal.

\item We add the SymFace loss to any generic face loss. The aggregated loss tends to aid the network in extracting the hidden information of asymmetry and helps to increase the inter-class variance among the classes.

\item We evaluate the added SymFace loss with various datasets (LFW, CFP-FP, CP-LFW, AgeDB, CA-LFW), and their results indicate the excellent potential of SymFace loss in the face recognition tasks.
\end{itemize}


\section{Literature Review}\label{section-literature-review}

The symmetrical behavior in face recognition solutions has been explored in the past. The paper \cite{FacialAsymmetry} explores how facial asymmetry affects facial recognition, focusing on expression variations, gender classification, and expression differentiation. The input images were computed using density difference (D-face) and edge orientation symmetry (S-face) to measure the asymmetry score. The author used principal component analysis to reduce the dimensions and performed the classification using the linear discriminant analysis (LDA) method. The other approach \cite{FaceLBPSYm} proposed a technique to enhance face recognition accuracy under occlusions and varying lighting. It integrates Local Binary Patterns (LBP) with multi-mirror symmetry to capture facial textures and leverage reflective properties. The process includes pre-processing images, calculating LBP histograms, and combining them with mirrored facial features for recognition using a nearest-neighbor classifier. The paper \cite{AsymmetryAgeFace} calculated the difference between the right and left half-face images in the input space and tried to classify facial images using the calculated difference along with other attributes of the person.

Advanced neural network methods in face recognition have been brought from diverse spheres of the domain. One area of particular importance is enhancing loss functions in face recognition. The classification task involves the evaluation of Softmax loss obtained from the images and their corresponding labels, as shown:

\begin{equation}
\mathcal{L}_{\textbf{CE}}(x_i) = -\log \frac{ \exp{ (W_{y_i} z_i ) } } { \sum^C_{ j=1 } \exp{ (W_{j} z_i ) } }
\label{ce_vector}
\end{equation}

Here, the $i$-th image sample $x_i$ is assigned label $y_i$ out of total classes $C$, and the embedding of $x_i$ is $z_i \in R^d$, where $d$ is the embedding size. The weight matrix is $W \in R^{d \times C}$, and the bias terms are set to zero.

Advancing from Softmax \cite{center_loss, NoisySoftmax, MarginalSoftmax}, then employing weight and embedding normalization \cite{NormFace, LMSoftmax, AMSoftmax}, led the research community into angular space

\begin{equation}
\mathcal{L}_{\textbf{CE}}(x_i) = -\log \frac{ \exp{( s (\cos {\theta_{y_i}} ) )} } { \sum^C_{ j=1 } \exp{( s (\cos {\theta_j} ))} }
\label{ce_cos}
\end{equation}

Here, $\theta$ is the angle between the feature $z_i$ and the weight $W_j$, and $s$ is the scaling factor in the angular space.

This forged a new frontier in face recognition. With this new approach, the margin played a crucial role in cosine space (Eq. \ref{cosface_fn}) \cite{cosface}, multiplication (Eq. \ref{sphereface_fn}) \cite{sface1}  and addition (Eq. \ref{arcface_fn}) \cite{arcface} of the margin in theta space instead of cosine space exhibiting outstanding results. 

\begin{equation}
\resizebox{0.5\textwidth}{!}{$L_{\textbf{CosFace}} = - \log \frac{ \exp{ (s \cos (\theta_{y_i})  - m ) } }{ \exp{ (s (\cos (\theta_{y_i})  - m )) } + \sum\limits^N_{j=1,j \neq y_i }\exp{(s \cos \theta_j})}$}
\label{cosface_fn}
\end{equation}

\begin{equation}
\resizebox{0.5\textwidth}{!}{$L_{\textbf{SphereFace}} = - \log \frac{ \exp{ (s \cos (\theta_{y_i} \times m) ) } }{ \exp{ (s \cos (\theta_{y_i} \times m) ) } + \sum\limits^N_{j=1,j \neq y_i }\exp{(s \cos \theta_j})}$}
\label{sphereface_fn}
\end{equation}

\begin{equation}
\resizebox{0.5\textwidth}{!}{$L_{\textbf{ArcFace}} = - \log \frac{ \exp{ (s \cos (\theta_{y_i} + m) ) } }{ \exp{ (s \cos (\theta_{y_i} + m) ) } + \sum\limits^N_{j=1,j \neq y_i }\exp{(s \cos \theta_j})}$}
\label{arcface_fn}
\end{equation}

In (Eq. \ref{cosface_fn} - Eq. \ref{arcface_fn}),  $\theta_{y_i}$ the most pivotal variable $m$ is the margin for adding the penalty.

The strategy used by \cite{CurricularFace} was training with lower margins for easy samples and higher margins for complex samples, emphasizing the practical nature of adaptiveness during the training cycles. AdaFace\cite{adaface} further emphasized the easy and hard samples based on image quality and utilized feature normalization for the quality assessment. We are simply introducing a loss function that can be added easily to any existing loss functions and help the combined face loss function be more powerful. Our proposed symmetrical loss function extends this approach by incorporating a natural phenomenon of facial symmetry into the cosine space. While traditional losses enhance embedding through angular margins, they do not explicitly leverage the inherent symmetry present in human faces. By integrating symmetry constraints into the cosine similarity framework, our method refines the embedding in a manner that aligns more closely with natural facial structures. 

Better results require a better network, and many network architectures are proposed to improve face recognition results. In the current scenario, mobile-based networks have become essential due to their widespread applications in autonomous vehicles, robotics, and unmanned aerial vehicles (UAVs). Tailoring the networks for edge devices necessitates considering their lower computational requirements \cite{Mobilenetv1, mobilenetv2, MobileNetV3, ghostnet, ShuffleNet_v2, PocketNet, MobileFaceforAgedb}, specifically regarding floating-point operations (FLOPs), and a reduction in parameters. While many networks have been proposed to address these requirements in face verification, \cite{MobileFaceNet, GhostFaceNet, shufflefacenet, mixfacenet}, achieving the anticipated results remains challenging. Some networks \cite{GhostFaceNet, EdgeFace} have managed to reduce FLOPs, thereby improving computational efficiency, but at the expense of higher numbers of parameters.
Conversely, networks \cite{PocketNet} with lower parameter counts often entail higher FLOPs, which is a trade-off dilemma. This underscores the need for innovative approaches to balance computational efficiency and model complexity. This research incorporates the proposed SymFace loss into existing lightweight face recognition architectures and presents significant advancements in their performance.

The ResNet50 and ResNet100 architectures have performed remarkably in various newly developed face recognition methods \cite{Unifying_Margin, uniface, UnitSFace, URL, VarGFaceNet}. But ResNet100 trained on MS1M-V2 has staggered with 99.82\% accuracy in LFW. Enhancing the loss function can elevate any network's capacity to discern and distinguish facial features more precisely. Other methods such as BroadFace \cite{BroadFace} optimizes face recognition by leveraging a linear classifier to consider a vast array of identities. The advancement in probabilistic face embeddings \cite{PFE} by Sphere Confidence Face \cite{SC_FR} computes the confidence learning in spherical space from the Euclidian space. Such refinement could potentially lead to significant improvements in accuracy and performance, ultimately advancing the capabilities of face recognition systems for various applications, including security, surveillance, biometrics, etc.

\section{Proposed Method}\label{section-methodology}
To leverage facial symmetry in developing a face recognition model, we introduce a facial frontness measure, denoted as $\rho$, to identify images that capture a frontal view of the face. We then split these images into two parts to generate separate embeddings, which are expected to be similar. This similarity is enforced using our additional facial symmetry loss, SymFace. This section describes our novel 3PSS (3-point symmetric split) algorithm, which measures facial frontness and performs image splitting. We will then explain how the SymFace loss is utilized during model training.

\subsection{3PSS}\label{section-3pss}

\subsubsection{Facial Frontness Measurement}
\begin{figure}[t]
    \centering
    \fbox{\includegraphics[width=0.75\linewidth]{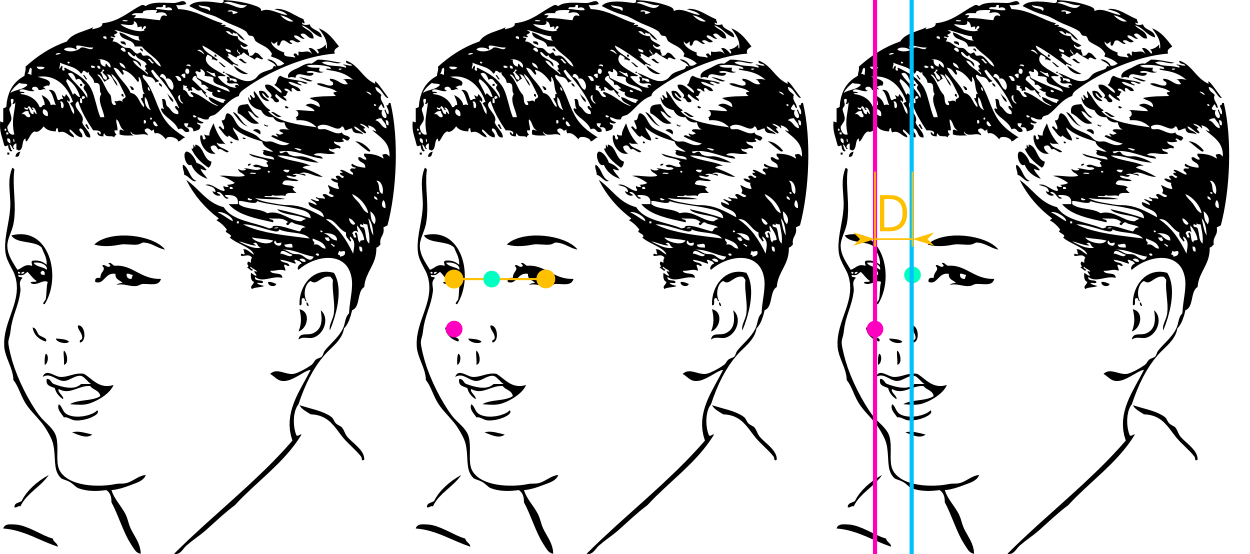}}
    \caption{Images being evaluated by the 3PSS}
    \caption*{Source: Pixabay, Free for use under the Pixabay Content License}
    \label{fig-cmp}
\end{figure}

We extract three facial landmarks, namely two eyes and a nose, using a pre-trained RetinaFace\cite{RetinaFace} model. We denote the x-coordinate of the landmarks of the left and right eyes as $e^l_x$ and $e^r_x$, respectively, and of the nose as $n_x$. The x-coordinate of the midpoint of the two eyes should be close to that of the nose to consider that the image has a frontal view of the face. A significant discrepancy in the two values suggests that either the face is tilted or not frontal, making it impractical to split the face. Thus, we calculate this discrepancy (denoted as $D$) as follows:
\begin{equation}
D = \Big|n_x - \frac{(e^l_x + e^r_x)}{2}\Big|, 
\end{equation}
which we use to compute what we call symmetric orientation coefficient($\rho$) in the following manner:

\begin{equation}\label{rho}
\rho = \frac{1}{1 + D^{2}}
\end{equation}

These steps can be visualized in Figure \ref{fig-cmp}, shown on a face-sketch along with landmarks of eyes (denoted in yellow dots) and nose(magenta dot). 

Note that we designate the $\rho$ value as zero for images where the landmark detector (RetinaFace) fails to detect landmarks. 

\subsubsection{Image Splitting}\label{section-data-preparation}
From Eq. \eqref{rho}, it's clear that the lower the discrepancy D, the higher the $\rho$ value. Thus, images with higher values of $\rho$ will be preferred for selecting frontal face images.

We use a threshold $\rho$ with $\tau$ (set as 0.2) to determine whether an image qualifies as symmetric or not, i.e., images with $\rho>\tau$ are considered "symmetric"; otherwise, and "asymmetric" otherwise. However, splitting every qualified image is undesirable because the network would never be trained on a full symmetric face image in such a case, and it would also increase the number of images more than required. So, only a fraction $p$ (set as 0.3) of these images are split in any epoch.

\begin{figure}[t]
    \centering
    \begin{subfigure}[b]{0.3\linewidth}
        \centering
        \fbox{\includegraphics[width=0.6\linewidth]{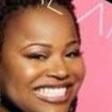}}
        \caption{Sample with score 0.01}
        \label{low-rho}
    \end{subfigure}
    \begin{subfigure}[b]{0.3\linewidth}
        \centering
        \fbox{\includegraphics[width=0.6\linewidth]{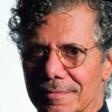}}
        \caption{Sample with score 0.38}
        \label{medium-rho}
    \end{subfigure}
    \begin{subfigure}[b]{0.3\linewidth}
        \centering
        \fbox{\includegraphics[width=0.6\linewidth]{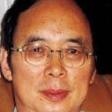}}
        \caption{Sample with score 0.5}
        \label{high-rho}
    \end{subfigure}
    \caption{A comparison of $\rho$ various images from the dataset}
    \label{fig:cmp-scores}
\end{figure}

Images in Fig.~\ref{fig:cmp-scores} show faces with various values of $\rho$. Fig.~\ref{low-rho} is considered asymmetrical due to its low $\rho$ value, whereas Fig.~\ref{medium-rho} and Fig.~\ref{high-rho} are considered symmetrical.

\begin{figure}
    \centering
    \fbox{\includegraphics[width=0.9\linewidth]{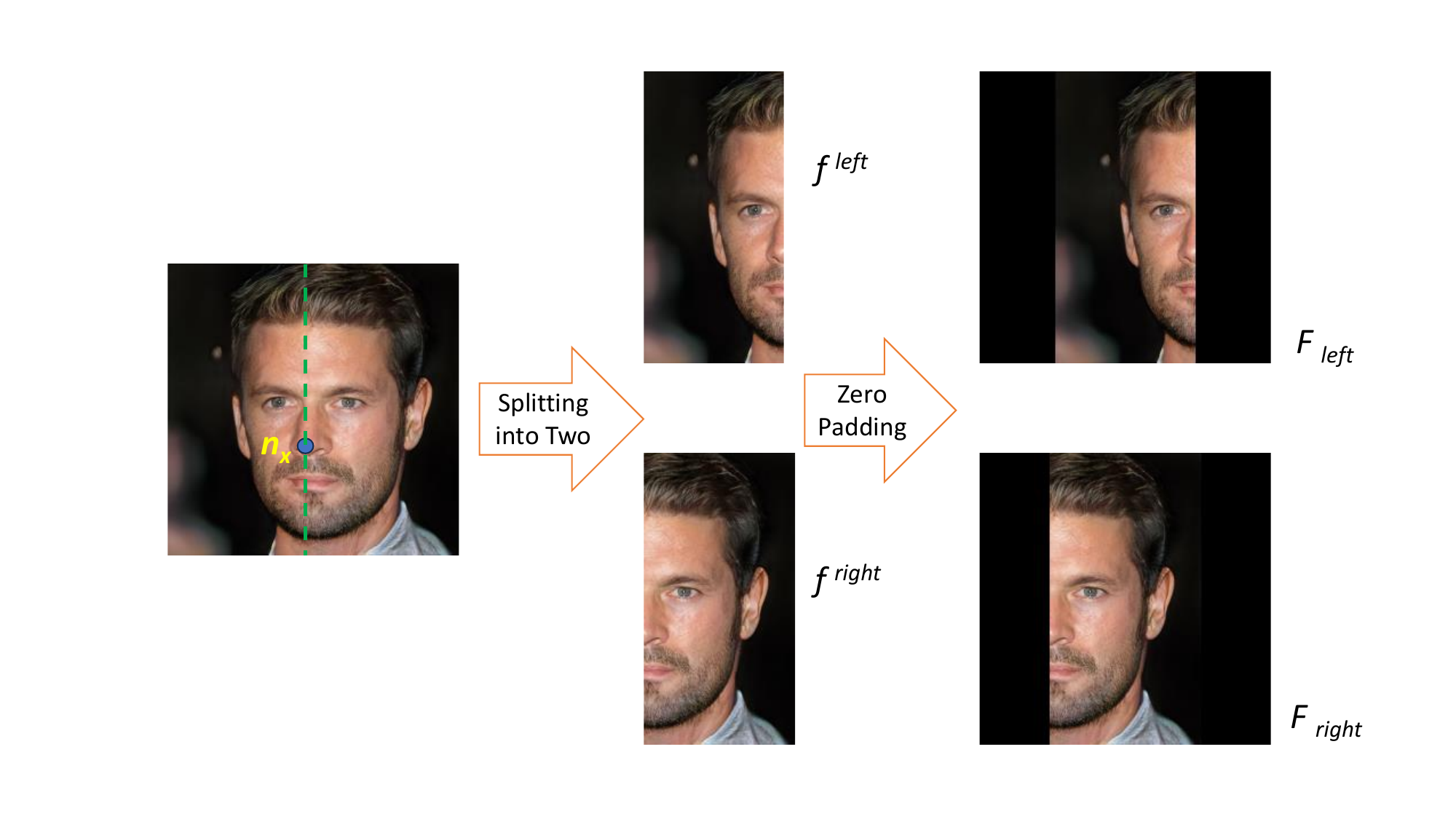}}
    \caption{Face Image Splitting Process}
    \label{face_split_k}
\end{figure}

For all images categorized as symmetric, we vertically split the face images into two halves: $f^{left}$ and $f^{right}$ for all images using $n_x$, the x-coordinate of nose landmark. While $f^{left}$ denotes the part of the image with columns up to $n_x$, $f^{right}$ denotes the remaining part. Both these parts are converted into an image of the same size as that of the original image by zero padding, such that they are at the center of the resulting images, $F_{left}$ and $F_{right}$, as shown in the Fig.~\ref{face_split_k}.  

\subsection{Training Procedure}\label{section-training-optimization}

\subsubsection{Training Samples}
In any epoch, with a fraction of randomly selected symmetrical images split, the number of training samples ($N$) in any epoch is increased as shown:

\begin{equation}
\begin{split}
N'&=N_{asym}+(1-p)N_{sym} + 2pN_{sym}\\
  &=N_{asym}+(1+p)N_{sym}\\
  &=N+pN_{sym}
\end{split}
\end{equation}
where $N_{sym}$ can be defined as follows:
\begin{equation}
N_{sym}=\sum\limits_{i=1}^{N}\delta(\rho_i>\tau),
\end{equation}
where $i$ is the index of the image in the original dataset. 
In any epoch, there will be two types of samples: full and half images. The full images are denoted as $x_i$, and the half images are denoted as $x_i^l$ or $x_i^r$ for left and right face images. Let $h_i\in\{0,1\}$ denote whether the full image $x_i$ has been split into $x_i^l$ and $x_i^r$ or not in a given epoch.

\begin{figure}
    \centering
    \fbox{\includegraphics[width=0.9\linewidth]{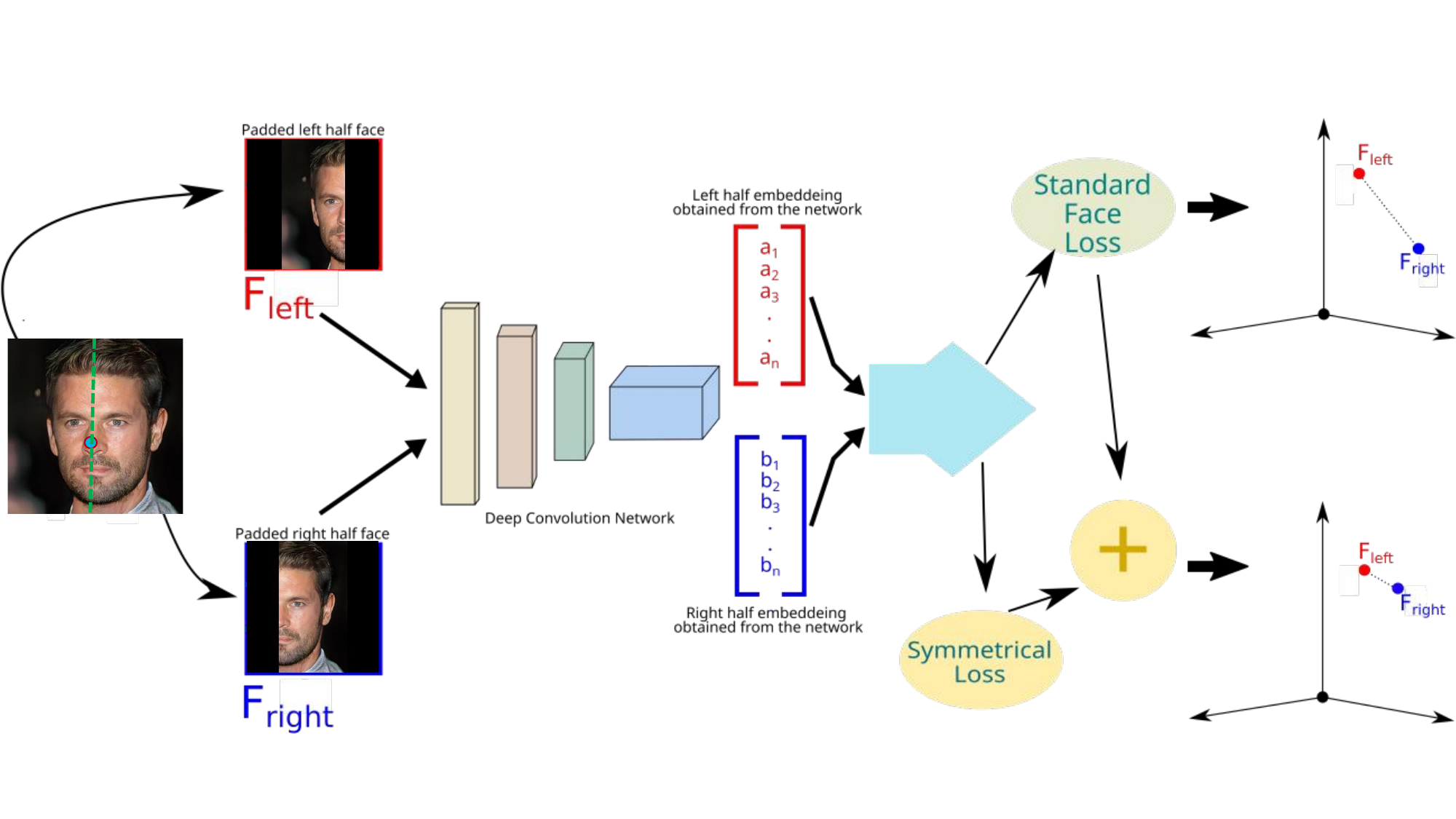}}
    \caption{Workflow of split samples}
    \label{fig:diagram}
\end{figure}

\subsubsection{SymFace Loss}
Our SymFace loss is defined as follows:

\begin{equation}
\mathcal{L}_\rho = \frac{1}{2pN_{sym}} \sum^N_{i=1} \rho_i\delta(h_i)||E(x_i^l) - E(x_i^r)||^2_2 
\label{loss_sym}
\end{equation}

Here, $E(x)$ denotes the output embedding of the face recognition network for image $x$. The idea is to apply our new loss only to the selected images out of the qualified ones for splitting. As shown in Fig.~\ref{fig:diagram}, this loss tries to bring the two embeddings closer.      

\subsubsection{Generic Face Loss} 
Let the standard loss function be denoted as $L_f(x,y)$ for a single sample, where $x$ is the image and $y$ is the image label. Our generic face loss can now be defined as follows:

\begin{equation}
\mathcal{L}_f=\frac{1}{N'}\sum\limits_{i=1}^{N}\delta(!h_i)L_f(x_i,y_i)+\delta(h_i)\Big(L_f(x^l_i,y_i)+L_f(x^r_i,y_i)\Big)
\end{equation}

Note that the denominator is $N'$, not $N$ because the loss is optimized for both full and half images.

\subsubsection{Total Loss}
Finally, we combine the two losses discussed to arrive at our final loss:
\begin{equation}
\mathcal{L}_{total}=\mathcal{L}_f+\mathcal{L}_\rho
\label{final-equation}
\end{equation}

\begin{table*}[htbp] 
    \centering
    \begin{tabular}{|c|c|c||c|c|}
        \hline
        \multicolumn{3}{|c||}{Configuration} & \multicolumn{2}{c|}{Validation Dataset} \\
        \hline
        \textbf{Model} & \textbf{Loss} & \textbf{Train Data} & \textbf{LFW} & \textbf{AgeDB}\\ \hline
        MobileFaceNet\cite{MobileFaceNet} & ArcFace & CASIA-WebFace  & 99.18 & \color{blue}\textbf{92.96} \\
        & ArcFace+SymFace & 112X96 & \color{red}\textbf{99.31} & 91.06 \\ \hline
        MobileFaceNet\cite{MobileFaceNet} & ArcFace & MS1MV2 & 99.55 & 96.07 \\
        & ArcFace+SymFace & & \color{red}\textbf{99.65} & \color{red}\textbf{96.08} \\ \hline
        ShuffleFaceNet x1.5 \cite{shufflefacenet}& ArcFace & MS1MV2 & 99.67 & \color{blue}\textbf{97.32} \\
        & ArcFace+SymFace & & \color{red}\textbf{99.73} & 96.71\\ \hline

    \end{tabular}
\vspace{5pt}
\caption{Verification performance (\%) on Lightweight Networks with the embedding size of 128}
\label{light-benchmarks} 
\end{table*}


\section{Experiments}\label{section-experiments}

\subsection{Datasets}\label{section-datasets}

We use the MS1MV2 dataset for the training cycle containing 5.8M facial images of 85K identities, and Webface \cite{WebFace} containing 4.2M facial images. The images in the dataset are first tagged with three facial landmarks using a pre-trained RetinaFace model \cite{RetinaFace}. These three landmark points, which include two eyes and one nose point, along with $\rho$ value against each image, are used as input along with the image data.

The validation cycle includes the following datasets: the LFW dataset \cite{LFW} containing 13,233 facial images of 5,749 people, the CFP-FP \cite{CFP-FP} dataset containing 7,000 facial images of 500 people, the CP-LFW \cite{CP-LFW} dataset contains 11,652 images of 5,749 people, the AgeDB \cite{AgeDB} dataset contains 16,488 facial images of 568 people, and the CA-LFW \cite{CA-LFW} dataset contains 12,174 facial images of 5,749 people.

\begin{table*}[htbp] 
    \centering
    \begin{tabular}{|c|c|c||c|c|c|c|c|}
        \hline
        \multicolumn{3}{|c||}{Configuration} & \multicolumn{5}{c|}{Validation Dataset} \\
        \hline
        \textbf{Model} & \textbf{Train Data} & \textbf{Loss} & \textbf{LFW} & \textbf{AgeDB} & \textbf{CA-LFW} & \textbf{CP-LFW} & \textbf{CFP-FP} \\
        \hline
        
        ResNet50 & MS1MV2 & AdaFace\cite{adaface} & 99.82 & 97.85 & 96.07 & 92.83 & 97.86 \\
        &  & AdaFace+SymFace & \color{red}\textbf{99.83} & \color{red}\textbf{97.89} & \color{red}\textbf{96.09} & \color{red}\textbf{93.29} & \color{red}\textbf{98.30} \\ \hline
        
        ResNet100 & MS1MV2 & AdaFace\cite{adaface} & 99.82 &  98.05 & 96.08 & \color{blue}\textbf{93.53} & \color{blue}\textbf{98.49} \\
        &  & AdaFace+SymFace & \color{red}\textbf{99.85} & \color{red}\textbf{98.08} & \color{red}\textbf{96.14} & 93.14 & 98.30 \\ \hline
        
        ResNet100 & MS1MV2 & CosFace\cite{adaface} & 99.81 & \color{blue}\textbf{98.11} & 95.76 &  92.28 & 98.12 \\
        &  & CosFace+SymFace & \color{red}\textbf{99.83} & 98.04 & \color{red}\textbf{96.08} & \color{red}\textbf{93.58} & \color{red}\textbf{98.40} \\ \hline
       
        ResNet100 & MS1MV2 & ArcFace & \color{blue}\textbf{99.83} & \color{blue}\textbf{98.28} & 95.45 & 92.08 & \color{blue}\textbf{98.27} \\
        &  & ArcFace+SymFace & 99.82 & 98.01 & \color{red}\textbf{95.99} & \color{red}\textbf{93.06} & 98.25\\ \hline
       
        ResNet50 &  WebFace4M & AdaFace\cite{adaface} & 99.78 & 97.78 & 95.98 & 94.17 & 98.97 \\
        &  & AdaFace+SymFace & \color{red}\textbf{99.83} & \color{red}\textbf{97.86}& \color{red}\textbf{96.01} & \color{red}\textbf{94.66}& \color{red}\textbf{99.01}\\ \hline
       
        ResNet100 & WebFace4M & CosFace\cite{WebFace} & 99.80 & 97.45 & \color{blue}\textbf{95.95} & 94.40 & \color{blue}\textbf{99.25} \\
        &  & CosFace+SymFace & \color{red}\textbf{99.83} & \color{red}\textbf{97.81} & 95.84 & \color{red}\textbf{94.46} & 99.15 \\ \hline
       
        ResNet100 & WebFace4M & ArcFace\cite{ArcRefWeb4M} & 99.83 & \color{blue}\textbf{98.28} & 95.45 & 92.08 & 98.27 \\
        &  & ArcFace+SymFace & \color{red}\textbf{99.83} & 97.61 & \color{red}\textbf{96.01} & \color{red}\textbf{94.21} & \color{red}\textbf{99.06}\\ \hline

        ResNet100 & WebFace4M & AdaFace\cite{adaface} & 99.80 & \color{blue}\textbf{97.90} & \color{blue}\textbf{96.05} & \color{blue}\textbf{94.63} & 99.17 \\
        &  & AdaFace+SymFace & \color{red}\textbf{99.85} & 97.83 & 95.93 & 94.53 & \color{red}\textbf{99.20}\\ \hline
    \end{tabular}
\vspace{5pt}
\caption{Verification performance (\%) on ResNet 50 and ResNet 100 with the embedding size 512}
\label{benchmarks} 
\end{table*}

\subsection{Experimental Settings}\label{section-experimental-settings}

In this experiment, the SymFace loss is used as an additional loss on top of existing loss functions ArcFace\cite{arcface} and AdaFace\cite{adaface}. The final combined loss is used during the training phase. We use different networks in this experiment to test the impact of SymFace loss. We use two lightweight networks, MobileFaceNet \cite{MobileFaceNet} with 0.99M parameters and ShufﬂeFaceNet \cite{shufflefacenet} with 2.6M parameters, as network backbones. We also use Resnet50 and ResNet100, as discussed in \cite{arcface}, as heavy network backbone. 

We use 3 A100 NVIDIA GPUs for lightweight networks and 8 A100 GPUs for heavy networks. As we make pairs in the dataset class, the batch size for lightweight networks is 75, while for heavy networks, it is 256. The network is fed with concatenated tensors, increasing the total batch sizes to 150 and 512 for lightweight and heavy networks, respectively. For lightweight networks and ResNet50, the network is trained up to 25 epochs. For ResNet100, the total number of epochs is set as 12. The initial learning rate is set as 0.1 for lightweight networks, and step scheduling is set at 17 (0.01). The initial learning rate for ResNet50 and ResNet100 is set at 0.01 and step scheduling at 10 (0.001). The SGD optimizer (momentum = 0.9) is used with weight decay as 4e-5 and 5e-4 for lightweight and heavy networks, respectively. Only for the last layer of MobileFaceNet the weight decay is set as 4e-4, as discussed in the paper \cite{MobileFaceNet}.

AdaFace and ArcFace loss functions are combined with SymFace loss as shown in (Eq. \ref{final-equation}). 128 and 512-sized embedding are used for lightweight and heavy networks, respectively. We normalize the image pixels by subtracting 127.5 and then dividing by 128. The horizontal flip is used during the training phase. For the lightweight networks with ArcFace+SymFace (ArcFace combined with SymFace) loss, the scale is set to 32, and the margin is 0.45. We do not add any augmentation (cropping, rescaling, and photometric jittering) in this research as introduced by \cite{adaface}. 

        
        

\subsection{Comparison Results}\label{section-comparison-results}

For fair comparisons of the results, we perform the experiments to the baseline configurations (network and loss function) \cite{MobileFaceNet, shufflefacenet, resnet, arcface, adaface}. The SymFace loss improves the discriminating power of ArcFace and AdaFace and pushes the network for better convergence.

For MobileFaceNet and ShuffleFaceNet, we compute ArcFace+SymFace loss with an embedding size of 128. The results surpass 66 \% of the times from the existing results as shown in Table \ref{light-benchmarks}. ResNet50 and ResNet100 are trained with additional SymFace loss and outperform the standard loss functions 70 \% of the time, and their corresponding SoTA results are populated in Table \ref{benchmarks}. The proposed loss function outperforms the LFW dataset most of the time. ResNet50 backbone achieves better results than its counterpart in most validation datasets; on the other hand, the ResNet100 backbone achieves better results in LFW and age-related datasets when trained on MS1M V2 but gives better results in side pose-related datasets when trained in the WebFace4M dataset. 

\begin{table*}[htbp] 
    \centering
    \begin{tabular}{|c|c|}
        \hline
        \textbf{Loss Function}  & \textbf{Inter-Class}\\ \hline
        ArcFace &  1.23 \\ \hline
        ArcFace + SymFace  & 3.09 \\ \hline
    \end{tabular}
\vspace{5pt}
\caption{Inter-class variance analysis on CASIA-WebFace on MobilefaceNet}
\label{Variance}
\end{table*}

\subsection{Ablation Study}\label{section-ablation}

\begin{figure}[t]
    \centering
    \fbox{\includegraphics[width=0.75\linewidth]{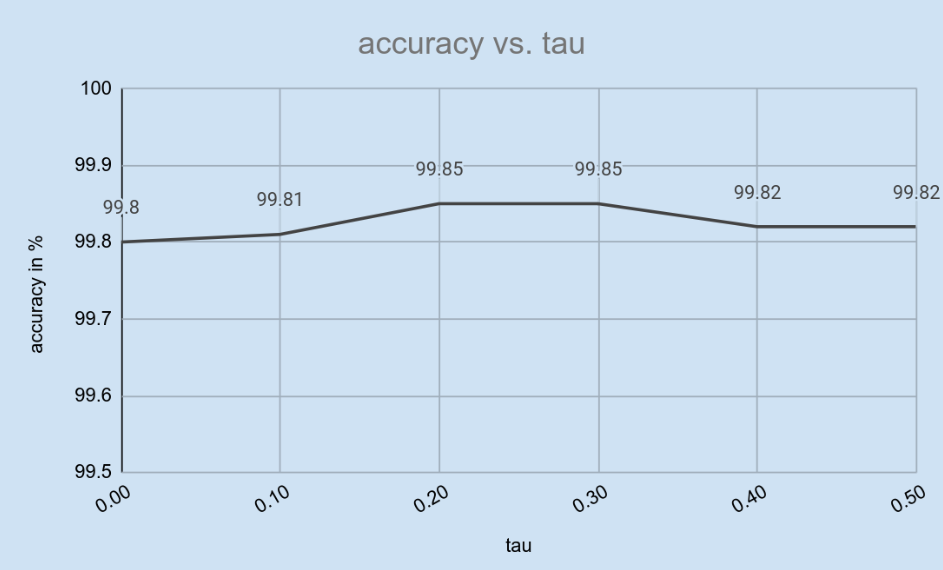}}
    \caption{Accuracy Vs Tau}
    \label{fig-cmp}
\end{figure}

\begin{table*}[htbp] 
    \centering
    \begin{tabular}{|c|c|}
        \hline
        \textbf{Method} & \textbf{LFW Accuracy} \\
        \hline
        AdaFace & 99.8  \\ \hline
        Split with AdaFace & 99.82\\ \hline
        Split with AdaFace+SymFace & 99.85 \\ \hline
    \end{tabular}
\vspace{5pt}
\caption{Comparison study of symmetrical loss With ResNet100 and AdaFace on WebFace4M dataset}
\label{ablation_study} 
\end{table*}

\begin{table*}[htbp] 
    \centering
    \begin{tabular}{|c|c|c|}
        \hline
        \textbf{Threshold} & \textbf{Cross-posed images \%} & \textbf{Cross-posed images \%}\\
        \textbf{value} & \textbf{(MS1MV2)} & \textbf{(WebFace4M)} \\ \hline
        0.05 & 38 \% & 42 \% \\ \hline
        0.1  & 48 \% & 54 \% \\ \hline
        0.2  & 59 \% & 66 \% \\ \hline
        0.3  & 66 \% & 73 \% \\ \hline
        0.4  & 74 \% & 78 \% \\ \hline
    \end{tabular}
\vspace{5pt}
\caption{Proportions of cross-posed or non-front aligned faces in the different datasets}
\label{cross_posed}
\end{table*}

 The network is trained with multiple values of $\tau$, and it is observed that very low (less than 0.1) or very high (0.3 or higher) values of $\tau$ do not enhance model performance (refer Fig.~\ref{fig-cmp}). The network reports better results for $\rho$ value in 0.2 $\pm$ 0.05 range. In our experiment, we find that feeding the network with the split augmentation but without symmetrical loss (only standard face loss) increases the model accuracy, but not up to the extent as with SymFace loss. When we train the ResNet100 network with split augmented images with standard loss only, it scores better at 99.82\% compared to the base network's accuracy of 99.8 \%. When the additional split images are fed to the ResNet100 network with additional symmetrical loss, we observe a better accuracy of 99.85\% (refer Table \ref{ablation_study}). Combining standard face loss with symmetrical face loss outperforms most of the validation datasets. We also apply the SymFace loss to the CASIA-WebFace of size $112X96$, and the accuracy on LFW increases from baseline results \cite{Mobileface} of 99.18\% to 99.31\% (Table \ref{light-benchmarks}). To explore the potential of the natural phenomenon of symmetry, we train the VarGFaceNet from scratch on MS1MV2 without knowledge distillation, with AdaFace loss, and get an accuracy of 99.76 on the LFW dataset- compared to 99.67 \% of the base model. 

 \section{Discussion}\label{section-discussion}
 
The results on the WebFace4M dataset in Table \ref{benchmarks} reveal that it outperforms pose variant datasets (CP-LFW and CFP-FP) compared to MS1M datasets. With the MS1MV2 dataset, the different loss functions scores in the range of 93.x \% and 98.0x \% in  CP-LFW and CFP-FP datasets, respectively but WebFace4M scores in the higher range of 94.x \% in CP-LFW and 99.x \% in CFP-FP datasets. This behavior is because the WebFace dataset contains quality pose variant images for lower values of $\rho$  (refer to Table \ref{cross_posed}). We analyze the improvement by the symmetrical loss for its discriminative capability of the network with better feature representation of distinct face images. The inter-class variance for the CASIA-WebFace dataset is analyzed, and the symmetrical loss is noted to enhance the inter-class distances among the classes, as shown in Table \ref{Variance}. The explanation for this behavior is obvious: the network is generally assigned a penalty for finding asymmetry. Thus, the network learns to extract the hidden features of asymmetric information to generate the output embedding and can differentiate different classes based on these hidden asymmetrical features.    

 \section{Conclusion}\label{section-conclusion}
This paper discusses SymFace loss applied as an additional loss over standard face losses. To use this loss, the facial landmarks are identified using a pre-trained RetinaFace model, based on which we define our metric of face orientation, namely symmetric orientation coefficient, expressed as $\rho$. This allows us to group facial images based on 2D face properties to extract symmetric features. We also present a customized training process, in which we pass pairs of images (along with labels and their corresponding $\rho$ values) and ensure that both types of facial images get processed by the total combined loss appropriately. The results support our hypothesis that such identification of symmetric features of the face can enhance the face recognition process. As the symmetry between the hemi faces is the core idea of this research, hence this method shows better results in LFW or age-based datasets (AgeDB, CA-LFW) but shows comparable results in the datasets focused on challenging side pose recognition (CP-LFW, CFP-FP). This further opens a new area of investigation to study pose-variant face samples in conjunction with symmetrical aspects.  

\section{Limitations}\label{section-limitations}
\subsection{Training Time}\label{section-training-time}
This approach can work only in datasets comprising at least 20-30 \% front-oriented face images. Another shortcoming is that this method employs additional training samples due to the face split process, which increases the training samples during the training phase. Though the inference time is always constant, we will explore further optimization of the SymFace loss in the future.

\subsection{Societal Impacts}\label{section-societal-impacts}

We stress that we do not condone or support the use of our work for mass surveillance and other repressive activities. We encourage the community and policymakers to establish clear regulations to prevent misuse. To mitigate this risk of False positives in security-related applications, we recommend deploying SymFace as part of a multi-factor authentication system, where facial recognition is one of several layers of security. In the context of our research, we utilize the MS1MV* training dataset \cite{MS-celeb-1M}, which is sourced from MS-Celeb, a dataset that its creator has officially withdrawn due to ethical concerns. Using MS1MV* allows us to conduct a fair and rigorous comparison of our findings against state-of-the-art methodologies in the field. We only use the MS1MV dataset to compare the results in our research.

\bibliographystyle{unsrt} 
\bibliography{main}


\end{document}